\documentclass[10pt,twocolumn,letterpaper]{article}

\usepackage[pagenumbers]{iccv} 
\definecolor{iccvblue}{rgb}{0.21,0.49,0.74}
\usepackage[pagebackref,breaklinks,colorlinks,allcolors=iccvblue]{hyperref}

\usepackage{graphicx}
\usepackage{amsmath}
\usepackage{amssymb}
\usepackage{booktabs}
\usepackage{multirow}
\usepackage{adjustbox}
\usepackage{wrapfig}
\usepackage{diagbox}
\usepackage{epsfig}
\usepackage{extpfeil}
\usepackage{makecell}
\usepackage{colortbl}
\usepackage{float}
\usepackage{wrapfig}
\usepackage{booktabs}
\usepackage{tabulary,overpic,xcolor}

\definecolor{cvprblue}{rgb}{0.21,0.49,0.74}

\usepackage[capitalize]{cleveref}
\usepackage{algorithm}
\usepackage{algorithmic}
\usepackage{bm}

\definecolor{color3}{rgb}{0.95,0.95,0.95}
\definecolor{color4}{rgb}{0.96,0.96,0.86}
\definecolor{color5}{rgb}{0.90,0.90,0.90}

\crefname{section}{Sec.}{Secs.}
\Crefname{section}{Section}{Sections}
\Crefname{table}{Table}{Tables}
\crefname{table}{Tab.}{Tabs.}

\title{Baking Gaussian Splatting into Diffusion Denoiser for Fast \\ and Scalable Single-stage Image-to-3D Generation and Reconstruction}

\author{Yuanhao Cai$^1$, He Zhang$^2$, Kai Zhang$^2$, Yixun Liang$^3$, \\ Mengwei Ren$^2$,  Fujun Luan$^2$, Qing Liu$^2$, Soo Ye Kim$^2$,  Jianming Zhang$^2$, \\ Zhifei Zhang$^2$,  Yuqian Zhou$^2$, Yulun Zhang$^4$, Xiaokang Yang$^4$, Zhe Lin$^2$, Alan Yuille$^1$ \\
$^{1}$ Johns Hopkins University, $^{2}$ Adobe Research, $^{3}$ HKUST, $^4$ Shanghai Jiao Tong University
}

\begin{document}
\twocolumn[{
\maketitle
\vspace{-7mm}\hspace{-2.5mm}
\centerline{
\includegraphics[width=0.98\linewidth,trim={4pt 4pt 4pt 4pt}]{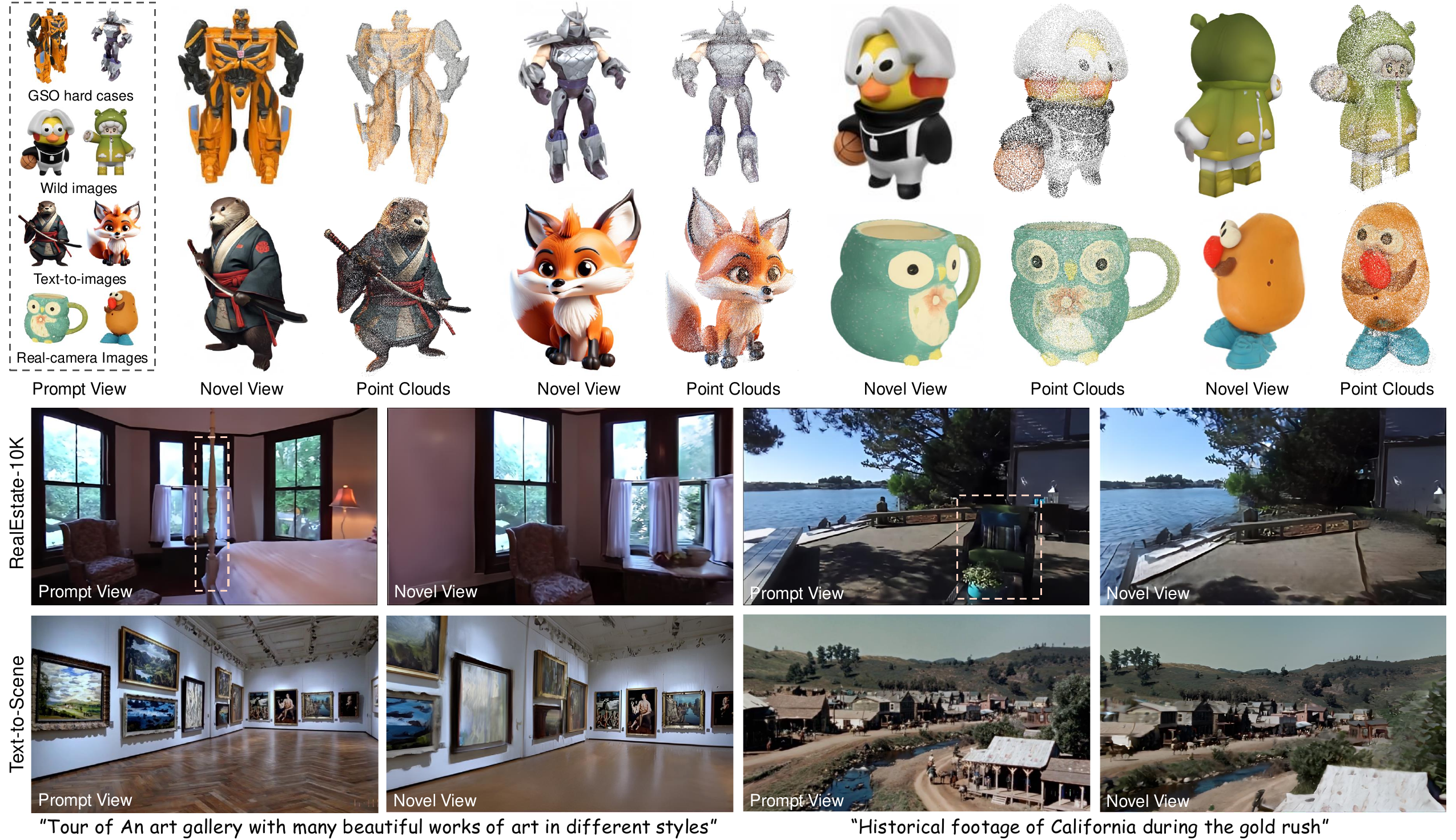}}
\vspace{-1mm}
\captionof{figure}
{
Single-view object generation (upper) and scene reconstruction (lower) results of our method. For single-view object generation, the prompt views are shown in the left dashed box. The generated novel views and 3D Gaussian point clouds are depicted on the right. For single-view scene reconstruction, our model can handle hard cases with occlusion and rotation, as illustrated in the dashed boxes of the third row. The prompt views of object and scene text-to-3D demos are generated by stable diffusion~\cite{stable_diffusion} and Sora~\cite{sora}, respectively.
}
\label{fig:teaser}
\vspace{3mm}
}]

\begin{abstract}
\vspace{-8mm}

\noindent Existing feedforward image-to-3D methods mainly rely on 2D multi-view  diffusion models that cannot guarantee 3D consistency. These methods easily collapse when changing the prompt view direction and mainly handle object-centric cases. In this paper, we propose a novel single-stage 3D diffusion model, DiffusionGS, for object generation and scene reconstruction from a single view. DiffusionGS directly outputs 3D Gaussian point clouds at each timestep to enforce view consistency and allow the model to generate robustly given prompt views of any directions, beyond object-centric inputs. Plus, to improve the capability and generality of DiffusionGS, we scale up 3D training data by developing a scene-object mixed training strategy. Experiments show that DiffusionGS yields improvements of 2.20 dB/23.25 and 1.34 dB/19.16 in PSNR/FID for objects and scenes than the state-of-the-art methods, without depth estimator. Plus, our method enjoys over 5$\times$ faster speed ($\sim$6s on an A100 GPU). Project page:   \url{https://caiyuanhao1998.github.io/project/DiffusionGS/}
\end{abstract}
\vspace{-1mm}

\begin{figure*}[t]
	\begin{center}
		\begin{tabular}[t]{c}  \hspace{-3mm}
\includegraphics[width=0.975\textwidth]{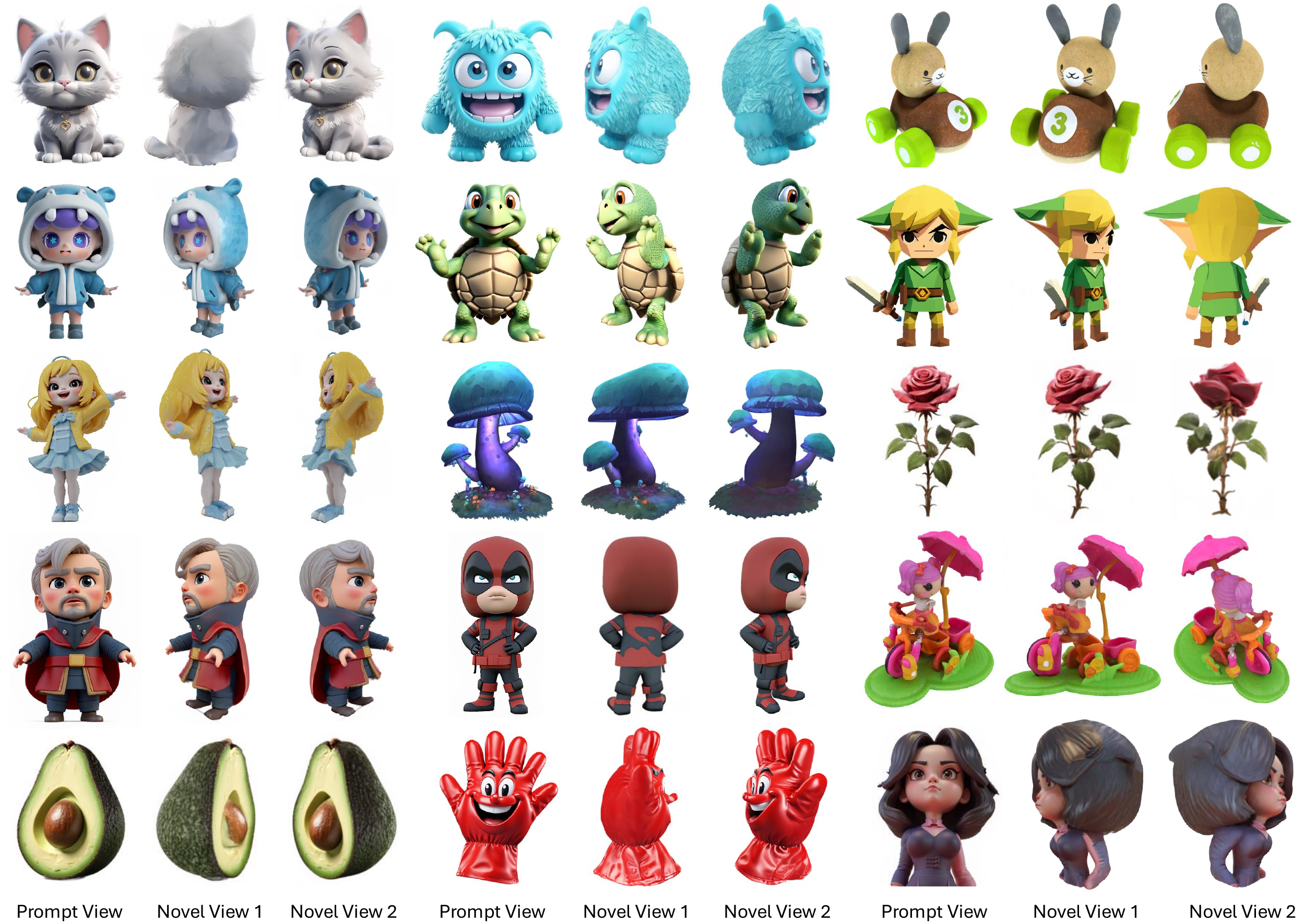}
		\end{tabular}
	\end{center}
	\vspace{-7mm}
	\caption{\small Single-view object generation results of our method on GSO~\cite{gso}, wild images, and text-to-images prompted by stable diffusion or FLUX. Our DiffusionGS can robustly handle hard cases with furry appearance, shadow, flat illustration, complex geometry, and specularity.}
	\label{fig:wild_object_demo}
	\vspace{-3mm}
\end{figure*}

\vspace{-5.8mm}
\section{Introduction}
\label{sec:intro}
\vspace{-1.2mm}
Image-to-3D task is important and challenging. It aims to reconstruct or generate a 3D representation of scenes or objects given only a single-view image. It has wide applications in AR/VR~\cite{ar_1}, film making~\cite{film_1}, robotics~\cite{robotics_1,robotics_2}, animation~\cite{animation_1,animation_2}, gaming~\cite{gaming_1}, and so on.

Existing feedforward image-to-3D methods are mainly two-stage~\cite{lgm,12345++,crm,instant3d}. They firstly adopt a 2D diffusion model to generate blocked multi-view images and secondly feed the multi-view images into a 3D reconstruction model. Without 3D model in the diffusion, these methods cannot enforce view consistency and easily collapse when the prompt view direction changes.  Another less studied technical route~\cite{renderdiffusion,viewsetdiff,dmv3d} is to train a 3D diffusion model with 2D rendering loss. Yet, these methods mainly rely on triplane neural radiance field (NeRF)~\cite{nerf}. The volume rendering of NeRF is time-consuming and the triplane resolution is limited, preventing the model from scaling up to larger scenes. In addition, current methods mainly study object-level generation using only object-centric datasets to train, which limits the model generalization ability and leaves larger-scale scene-level cases less explored. Although few recent works~\cite{vistadream, flash3d} study single-view scene reconstruction, they rely on monocular depth estimators and easily collapse under severe occlusion or large viewpoint changes.

\begin{figure*}[t]
	\begin{center}
		\begin{tabular}[t]{c}  \hspace{-3mm}
			\includegraphics[width=1\textwidth]{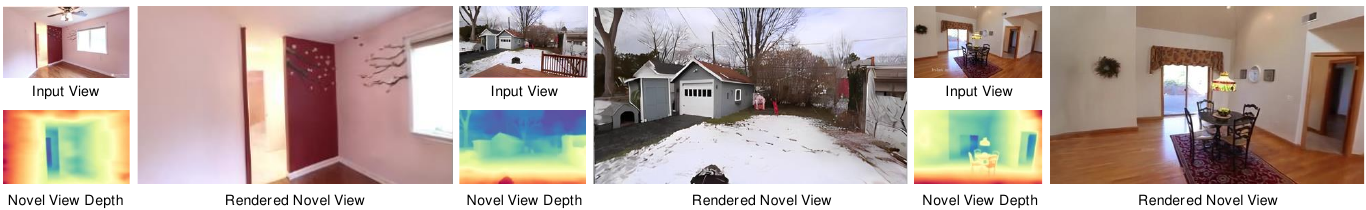}
		\end{tabular}
	\end{center}
	\vspace{-7.2mm}
	\caption{\small Single-view scene reconstruction of our method on indoor and outdoor scenes. The depth maps are rendered by GS point clouds. }
	\label{fig:scene_demo}
	\vspace{-3.8mm}
\end{figure*}

To address these issues, we propose a novel single-stage 3D Gaussian Splatting (3DGS)~\cite{3dgs} based diffusion model, DiffusionGS, 
for 3D object generation and scene reconstruction from a single view. Our DiffusionGS enforces 3D consistency of the generated contents by predicting multi-view pixel-aligned  Gaussian primitives in every timestep. With the highly parallel rasterization and scalable imaging range,  DiffusionGS enjoys a fast inference speed of $\sim$6 seconds and can be easily applied to large scenes. Unlike previous single-view scene reconstruction methods~\cite{pixelnerf,vistadream,flash3d} that predict 3D representation only from the given view, DiffusionGS generates other views along the camera trajectory to predict more refined and structured Gaussian point clouds. Thus, our method can better perceive the geometry to reconstruct the scene \textbf{without using depth estimator}. As our goal is to build a general and large-scale 3D generation model, it is critical to fully exploit existing 3D scene and object data. Yet, directly training with scene and object data may lead to non-convergence because of the large domain discrepancy. Thus, we propose a scene-object mixed training strategy to handle this problem and learn a general prior of geometry and texture. Our mixed training strategy adapts DiffusionGS to both object and scene datasets by controlling the distribution of the selected views, camera condition, Gaussian point clouds, and imaging depths. In particular, we notice previous camera conditioning method Pl\"ucker coordinate~\cite{plucker_coordinate} shows limitations in capturing depth and 3D geometry. Hence, we design a new camera conditioning method, Reference-Point Pl\"ucker Coordinates (RPPC), that encodes the point on each ray closest to the origin of the world coordinate system to help DiffusionGS better perceive the depth and 3D geometry across scene and object data. Finally, the model is further finetuned on object or scene data, respectively, to boost the performance.

Our contributions can be summarized as follows:

\begin{itemize}
	\vspace{0mm}
        \item We propose a novel framework, DiffusionGS, for 3D object generation and scene reconstruction from single view. 
	\vspace{-3mm}
	\item We design a scene-object mixed training strategy to learn a more general prior from both 3D object and scene data.
	\vspace{-3mm}
        \item We customize a new camera pose conditioning method, RPPC, to better perceive the relative depth and geometry.
	\vspace{-3mm}
	\item Our method outperforms SOTA single-view object generation and scene reconstruction methods by 2.20 dB/23.25 and 1.34 dB/19.16 in PSNR/FID score, while enjoying a fast inference speed of $\sim$6s on a single A100 GPU.
\end{itemize}

\vspace{-1.5mm}
\section{Related Work}
\label{sec:rel_work}
\vspace{-1mm}
\subsection{Diffusion Models for Image-to-3D Generation}
\vspace{-0.5mm}
Diffusion models~\cite{ddpm,score_dm,ddim,stable_diffusion} are proposed for image generation and recently have been applied to 3D generation, which can be divided into four categories. The first category~\cite{point-e, shape-e, direct_3d_1, direct_3d_2, direct_3d_3, diffgs, gaussiancube} uses direct supervision on 3D models such as point clouds or meshes, which are hard to obtain in practice. The second kind of methods~\cite{dreamgaussian,gaussiandreamer,prolificdreamer,dreamreward,dreambooth3d} use SDS loss~\cite{sds} to distill a 3D model from a 2D diffusion. Yet, these methods require a time-consuming per-asset optimization. The third category~\cite{zero123, view_diff_1, shi2023mvdream, gu2023nerfdiff, liu2023syncdreamer, wonder3d, multidiff} adds the camera poses as the input condition to finetune a 2D diffusion model to render fixed novel views. Yet, these methods cannot guarantee 3D consistency and easily collapse when prompt view direction changes. The last category~\cite{dmv3d,renderdiffusion,viewsetdiff,ddibr} trains a 3D diffusion model with 2D rendering loss. However, these methods mainly based on triplane-NeRF suffer from the limited resolution of triplane and slow speed of volume rendering. 


\vspace{-1mm}
\subsection{Gaussian Splatting}
\vspace{-0.5mm}
3DGS~\cite{3dgs} uses millions of Gaussian ellipsoid point clouds to represent objects or scenes and render views with rasterization. It achieves success in 3D/4D reconstruction~\cite{mip-gs, x_gaussian, FSGS, dynamic1, dynamic2, dynamic3, lightgaussian, lsm, cogs, mvsplat, geogaussian}, generation~\cite{gaussiandreamer,luciddreamer,humangaussian,hugs,gauhuman,gaussiancube,diffgs, gaussianeditor, gsd}, inverse rendering~\cite{InverseRendering1,InverseRendering2,InverseRendering3}, SLAM~\cite{slam1,slam3, slam4}, \emph{etc.} For instance, Flash3D~\cite{flash3d} and VistaDream~\cite{vistadream} use a monocular depth estimator to predict GS point clouds for single-view scene reconstruction. However, these methods show limitations in handling scenes with severe occlusion and large viewpoint changes. 

\begin{figure*}[t]
	\begin{center}
		\begin{tabular}[t]{c}  \hspace{-2.4mm}
			\includegraphics[width=1.0\textwidth]{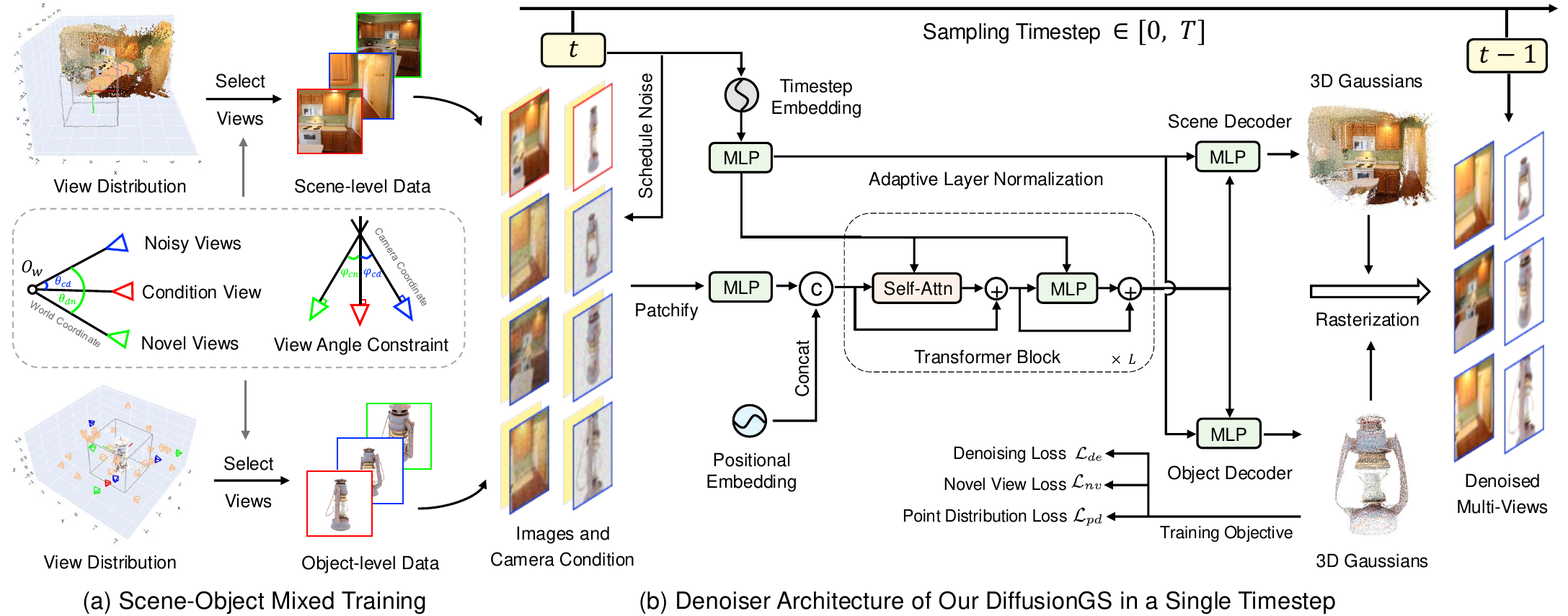}
		\end{tabular}
	\end{center}
	\vspace{-7mm}
	\caption{\small Pipeline. (a) When selecting the data for our scene-object mixed training, we impose two angle constraints on the positions and orientations of viewpoint vectors to guarantee the training convergence. (b) The denoiser architecture of DiffusionGS in a single timestep.}
	\label{fig:pipeline}
	\vspace{-3mm}
\end{figure*}

\vspace{-2mm}
\section{Method}
\label{sec:method}
\vspace{-1mm}

Fig.~\ref{fig:pipeline} depicts the pipeline of our method.  Fig.~\ref{fig:pipeline} (a) shows the scene-object mixed training. 
For each scene or object, we pick up a view as the condition, $N$ views as the noisy views to be denoised, and $M$ novel views for supervision. Then in Fig.~\ref{fig:pipeline} (b), the clean and noisy views are fed into our DiffusionGS to predicts per-pixel 3D Gaussian primitives. 

\vspace{-1mm}
\subsection{DiffusionGS}
\vspace{-0.5mm}

\noindent \textbf{Preliminary of Diffusion.} We first review denoising diffusion probabilistic model (DDPM)~\cite{ddpm}. In the forward noising process, DDPM transforms the real data distribution $x_0 \sim q(x)$ to standard normal distribution $\mathcal{N}(0, \mathbf{I})$ by gradually applying noise to the real data $x_0: q(x_t | x_0) = \mathcal{N}(x_t; \sqrt{\bar{\alpha}_t}x_0, (1-\bar{\alpha_t})\mathbf{I})$ at every timestep $t \in [0, T]$, where $\bar{\alpha_t}$ are pre-scheduled hyper-parameters. Then $x_t$ is sampled by   $x_t = \bar{\alpha_t} x_0 + \sqrt{1-\bar{\alpha_t}}\epsilon_t$, where $\epsilon_t \sim \mathcal{N}(0, \mathbf{I})$. The denoising process reverses the forward process by gradually using a neural network to predict $\epsilon_t$. Similarly, 2D multi-view diffusion~\cite{zero123,shi2023mvdream,lgm,12345++} generates novel views by denoising images or latents at multiple viewpoints. However, these 2D diffusions do not have 3D models, thus suffering from view misalignment and easily collapsing when the prompt view direction changes. We solve these problems by baking 3D Gaussians into the diffusion denoiser.

\noindent\textbf{Our 3D Diffusion.} Different from the normal diffusion model that predicts noise, our DiffusionGS aims to recover clean 3D Gaussian point clouds. Thus, we design the denoiser to directly predict pixel-aligned 3D Gaussians~\cite{splatter_image} and be supervised at clean 2D multi-view  renderings. 



As shown in Fig.~\ref{fig:pipeline} (b), the input of DiffusionGS in the training phase are one clean condition view $\mathbf{x}_{con} \in \mathbb{R}^{H\times W \times 3}$ and $N$ noisy views $\mathcal{X}_t = \{\mathbf{x}_t^{(1)}, \mathbf{x}_t^{(2)}, \cdots, \mathbf{x}_t^{(N)}\}$ concatenated with viewpoint conditions $\mathbf{v}_{con} \in \mathbb{R}^{H\times W \times 6}$ and $\mathcal{V} = \{\mathbf{v}^{(1)}, \mathbf{v}^{(2)}, \cdots, \mathbf{v}^{(N)}\}$. Denote the clean counterparts of the noisy views as $\mathcal{X}_0 = \{\mathbf{x}_0^{(1)}, \mathbf{x}_0^{(2)}, \cdots, \mathbf{x}_0^{(N)}\}$. The forward diffusion process adds noise to each view as
\vspace{-1.5mm}
\begin{equation}
\small
    \mathbf{x}_t^{(i)} = \bar{\alpha_t} \mathbf{x}_0^{(i)} + \sqrt{1-\bar{\alpha_t}}\epsilon_t^{(i)},
    \label{eq:q_sample}
\vspace{-1.5mm}
\end{equation}
where $\epsilon_t^{(i)} \sim \mathcal{N}(0, \mathbf{I})$ and $i = 1, 2, \cdots, N$. Then in each timestep $t$, the denoiser $\theta$ predicts the 3D Gaussians $\mathcal{G}_{\theta}$ to enforce view consistency. As the number of original 3D Gaussians is not a constant, we adopt the pixel-aligned 3D Gaussians~\cite{splatter_image} as the output, whose number is fixed. The predicted 3D Gaussians $\mathcal{G}_{\theta}$ is formulated as
\vspace{-0.5mm}
\begin{equation}
\small
    \mathcal{G}_{\theta} (\mathcal{X}_t | \mathbf{x}_{con}, \mathbf{v}_{con}, t, \mathcal{V}) = \{G_t^{(k)}(\bm{\mu}_t^{(k)}, \mathbf{\Sigma}_t^{(k)}, \alpha_t^{(k)}, \bm{c}_t^{(k)})\},
    \label{eq:denoiser}
\vspace{-0.5mm}
\end{equation}
where $1 \leq k \leq N_g$ and $N_g = (N+1)HW$ is the number of per-pixel Gaussian $G_t^{(k)}$. $H$ and $W$ are the height and width of the image. Each $G_t^{(k)}$ contains a center position $\bm{\mu}_t^{(k)} \in \mathbb{R}^3$, a covariance $\mathbf{\Sigma}_t^{(k)} \in \mathbb{R}^{3\times3}$ controlling its shape, an opacity $\alpha_t^{(k)} \in \mathbb{R}$ characterizing the transmittance, and an RGB color $\bm{c}_t^{(k)} \in \mathbb{R}^3$.  Specifically, $\bm{\mu}_t^{(k)} = \bm{o}^{(k)} + u_t^{(k)} \bm{d}^{(k)}$. $\bm{o}^{(k)}$ and $\bm{d}^{(k)}$ are the origin and direction of the $k$-th pixel-aligned ray. The distance $u_t^{(k)}$ is parameterized by 
\vspace{-1mm}
\begin{equation}
\small
    u_t^{(k)} =  w_t^{(k)} u_{near} + (1 - w_t^{(k)}) u_{far},
\vspace{-1mm}
\label{eq:distance}
\end{equation}
where $u_{near}$ and $u_{far}$ are the nearest and farthest distances. $w_t^{(k)} \in \mathbb{R}$ is the weight to control $u_{t}^{(k)}$. $\mathbf{\Sigma}_t^{(k)}$ is parameterized by a rotation matrix $\mathbf{R}_t^{(k)}$ and a scaling matrix  $\mathbf{S}_t^{(k)}$. $w_t^{(k)}, \mathbf{R}_t^{(k)}, \mathbf{S}_t^{(k)}, \alpha_t^{(k)},$ and $ \bm{c}_t^{(k)}$ are directly extracted from the merged per-pixel Gaussian maps by splitting channels.

\noindent\textbf{Denoiser Architecture.} As shown in Fig.~\ref{fig:pipeline} (b), the input images concatenated with the viewpoint conditions are patchified, linearly projected, and then concatenated with a positional embedding to derive the input tokens of the Transformer backbone, which consists of $L$ blocks. Each block contains a multi-head self-attention (MSA), an MLP, and two layer normalization (LN). Eventually, the output tokens  are fed into the Gaussian decoder to be linearly projected and then unpatchified into per-pixel Gaussian maps $\hat{\mathcal{H}} = \{\hat{\mathbf{H}}_{con}, \hat{\mathbf{H}}^{(1)}, \cdots, \hat{\mathbf{H}}^{(N)}\}$, where $\hat{\mathbf{H}}_{con}$ and $\hat{\mathbf{H}}^{(i)} \in \mathbb{R}^{H\times W\times 14}$. Then $N+1$ Gaussian maps are merged into the Gaussian point clouds $\mathcal{G}_{\theta}$ in Eq.~\eqref{eq:denoiser}. The timestep condition controls the Transformer block and Gaussian decorder through the adaptive layer normalization mechanism~\cite{dit}.

\noindent\textbf{Gaussian Rendering.} As the ground truth of $G_t^{(k)}$ is not available, we use the 2D renderings to supervise $\mathcal{G}_{\theta}$. To this end, we formulate DiffusionGS to a multi-view diffusion model. As aforementioned, 2D diffusion usually predicts the noise $\epsilon_t$. Yet, noisy Gaussian point clouds do not have texture information and may degrade view consistency. To derive clean and complete 3D Gaussians, DiffusionGS is $x_0$-prediction instead of $\epsilon$-prediction. The denoised multi-view images $\hat{\mathcal{X}}_{(0,t)} = \{\hat{\mathbf{x}}_{(0,t)}^{(1)}, \hat{\mathbf{x}}_{(0,t)}^{(2)}, \cdots, \hat{\mathbf{x}}_{(0,t)}^{(N)}\}$ are rendered by the differentiable rasterization function $F_r$ as
\vspace{-1mm}
\begin{equation}
\small
    \hat{\mathbf{x}}_{(0,t)}^{(i)} = F_r(\mathbf{M}_{ext}^{(i)}, \mathbf{M}_{int}^{(i)}, \mathcal{G}_{\theta}(\mathcal{X}_t | \mathbf{x}_{con}, \mathbf{v}_{con}, t, \mathcal{V})),
\vspace{-1mm}
\end{equation}
where $1 \leq i \leq N$. $\mathbf{M}_{ext}^{(i)}$ and $\mathbf{M}_{int}^{(i)}$ denote the extrinsic matrix and intrinsic matrix of the viewpoint $\mathbf{c}^{(i)}$.

For each $G_t^{(k)}$ at viewpoint $\mathbf{c}^{(i)}$, the rasterization projects its 3D covariance $\mathbf{\Sigma}_{t}^{(k)}$ from the world coordinate system to $\mathbf{\Sigma'}_t^{(k, i)} \in \mathbb{R}^{3\times 3}$ in the camera coordinate system as
\vspace{-1mm}
\begin{equation}
\small
	\mathbf{\bf \Sigma'}_t^{(k, i)} = \mathbf{J}_t^{(i)} \mathbf{W}_t^{(i)} \mathbf{\Sigma}_t^{(k)} {\mathbf{W}_t^{(i)}}^\top {\mathbf{J}_t^{(i)}}^\top,
\vspace{-1.8mm}
\end{equation}
where $\mathbf{J}_t^{(i)} \in \mathbb{R}^{3\times 3}$ is the Jacobian matrix of the affine approximation of the projective transformation. $\mathbf{W}_t^{(i)} \in \mathbb{R}^{3\times 3}$ is the viewing transformation. 
The 2D projection is divided into non-overlapping tiles. The 3D Gaussians are assigned to the tiles where their 2D projections cover. For each tile, the assigned 3D Gaussians are sorted according to the view space depth. Then the RGB value at pixel $(m, n)$ is derived by blending $\mathcal{N}$ ordered points overlapping the pixel as
\vspace{-1mm}
\begin{equation}
\small
\begin{aligned}
	\hat{\mathbf{x}}_{(0,t)}^{(i)}(m, n) = \sum_{j \in \mathcal{N}} \bm{c}_t^{(j)}~\sigma_t^{(j)}  \prod_{l=1}^{j-1}(1-\sigma_t^{(l)}),
\end{aligned}
\vspace{-1mm}
\end{equation}
\begin{figure}[t]
	\begin{center}
		\begin{tabular}[t]{c}  \hspace{-1mm}
	\includegraphics[width=0.45\textwidth]{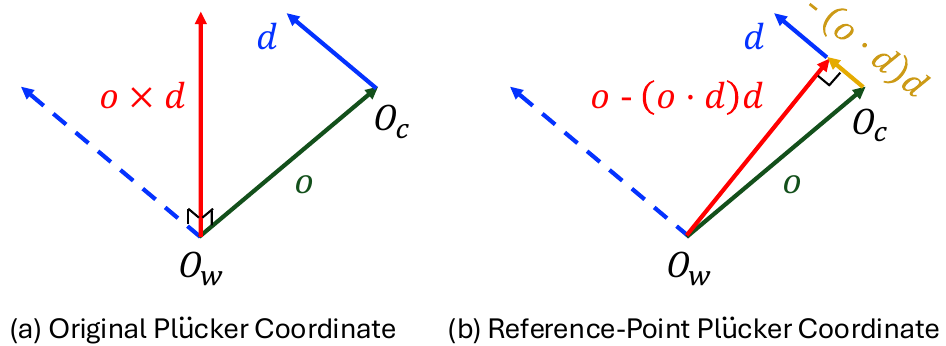}
		\end{tabular}
	\end{center}
	\vspace{-6mm}
	\caption{\small Pl\"ucker ray \emph{vs.} Reference-Point Pl\"ucker Coordinate.}
	\label{fig:rppc}
	\vspace{-5mm}
\end{figure}
where $\sigma_t^{(l)} = \alpha_t^{(l)} P(\mathbf{z}_t^{(l)} | \bm{\mu}_t^{(l)}, \mathrm{\bf \Sigma}_t^{(l)})$,  $\mathbf{z}_t^{(l)}$ is the $l$-th intersection 3D point, and  $P(\mathbf{z}_t^{(l)} | \bm{\mu}_t^{(l)}, \mathrm{\bf \Sigma}_t^{(l)})$ is the possibility  of the corresponding 3D Gaussian distribution at $\mathbf{z}_t^{(l)}$.

Then we use the weighted sum, controlled by $\lambda$, of $\mathcal{L}_2$ loss and VGG-19~\cite{vgg} perceptual loss $\mathcal{L}_{\text{VGG}}$ between the multi-view predicted images $\hat{\mathcal{X}}_{(0,t)}$ and ground truth ${\mathcal{X}}_{0}$ as the denoising loss $\mathcal{L}_{de}$ to supervise the 3D Gaussians $\mathcal{G}_{\theta}$ as
\vspace{-0.8mm}
\begin{equation}
\small
    \mathcal{L}_{de} = \mathcal{L}_2 (\hat{\mathcal{X}}_{(0,t)}, {\mathcal{X}}_{0}) + \lambda \cdot \mathcal{L}_{\text{VGG}} (\hat{\mathcal{X}}_{(0,t)}, {\mathcal{X}}_{0}).
\vspace{-0.8mm}
\label{eq:denoise_loss}
\end{equation}
In the testing phase, our DiffusionGS randomly samples noise from standard normal distribution at timestep $T$ and then gradually denoise it step by step. The predicted $\hat{\mathcal{X}}_{(0,t)}$ at each timestep $t$ is fed into the next timestep $t-1$ to replace $\mathcal{X}_0$ in Eq.~\eqref{eq:q_sample} for sampling $\mathcal{X}_{t-1}$ at each noisy view as
\vspace{-2mm}
\begin{equation}
\small
    \mathbf{x}_{t-1}^{(i)} = \bar{\alpha}_{t-1} \mathbf{\hat{x}}_{(0, t)}^{(i)} + \sqrt{1-\bar{\alpha}_{t-1}}\epsilon_{t-1}^{(i)}.
    \label{eq:diff_sample}
\vspace{0mm}
\end{equation}
Then we use 30-step DDIM~\cite{ddim} to facilitate the sampling speed by skipping some timesteps. One clean image and relative poses are input for inference. Finally, the generated $\mathcal{G}_{\theta}$ at $t = 0$ in Eq.~\eqref{eq:denoiser} can be used to render novel views.

\begin{figure*}[t]
	\begin{center}
		\begin{tabular}[t]{c}  \hspace{-4mm}
			\includegraphics[width=0.944\textwidth]{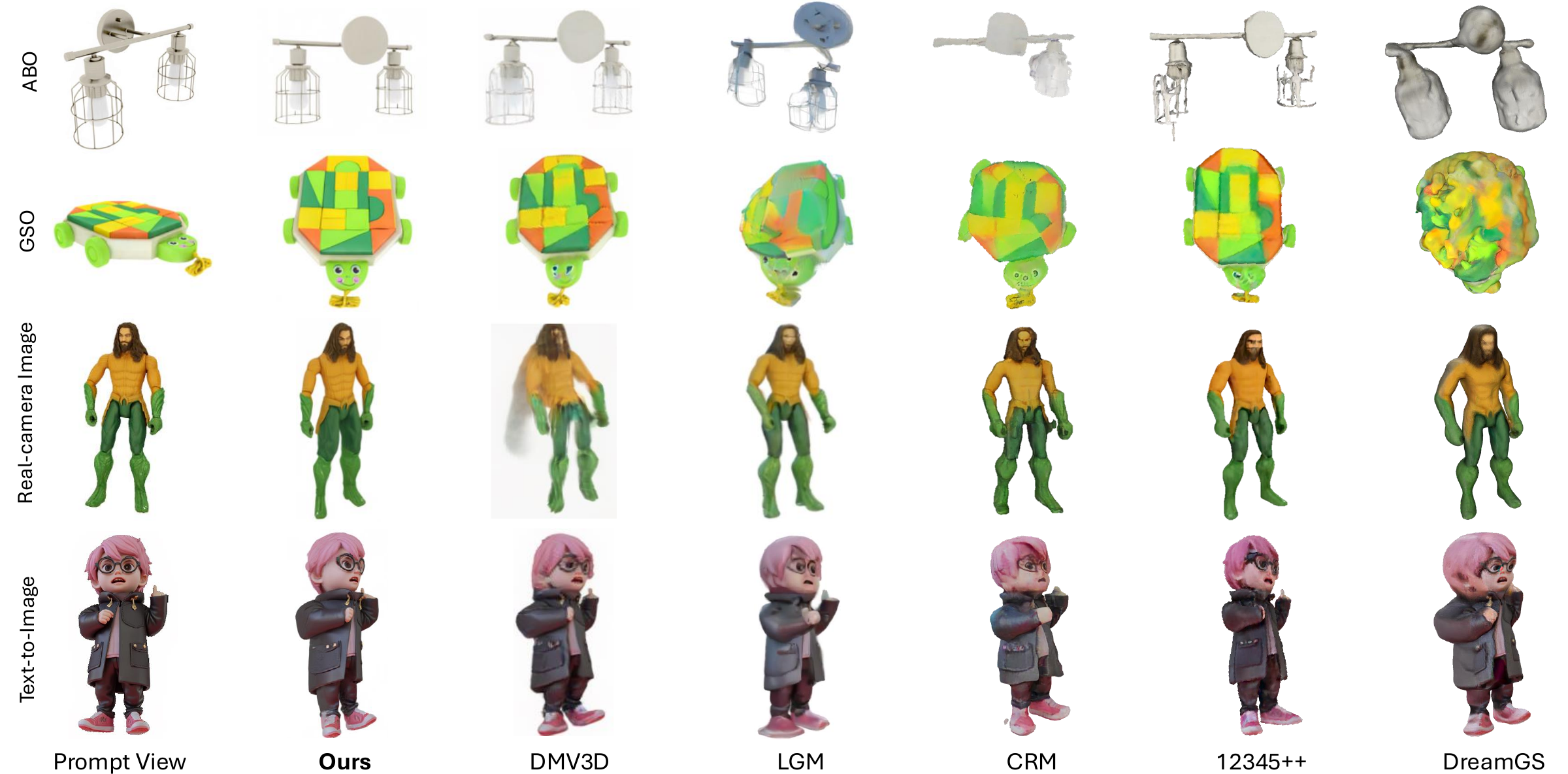}
		\end{tabular}
	\end{center}
	\vspace{-7mm}
	\caption{\small Visual comparison of single-view object generation on ABO, GSO, real-camera image, and text-to-image prompted by FLUX. Our method can generate more fine-grained details with accurate geometry from prompt views of any directions. Zoom in for a better view.}
	\label{fig:object_compare}
	\vspace{-4mm}
\end{figure*}

\vspace{-1mm}
\subsection{Scene-Object Mixed Training Strategy}
\vspace{-1mm}
Existing 3D training data is relatively scarce and the cost of data annotation is expensive. Especially for the scene-level datasets, there are only $\sim$90K training samples~\cite{realestate10k,dl3dv} and most of them only cover small viewpoint changes, which are not sufficient to learn strong geometry representations. Besides, the majority of object-level training data is synthetic~\cite{objaverse}. As a result, models~\cite{dmv3d,instant3d,shape-e,point-e,lgm} trained on object-level data often generate unrealistic textures, limiting the practice on real-camera images. 
To improve the capacity and generalization ability of our DiffusionGS, it is critical to make full use of both object and scene data.

Yet,  directly training 3D diffusion models with both object and scene datasets may introduce artifacts or lead to non-convergence because of the large domain discrepancy. As shown in the lower part of Fig.~\ref{fig:pipeline} (a), object-level datasets~\cite{objaverse} usually have an object in the central position without background and the camera rotates around this central object to capture multi-view images. The imaging range and depth are limited. In contrast, as depicted in the upper part of Fig.~\ref{fig:pipeline} (a), scene-level datasets have more dense image representations instead of blank background. The imaging range and depth are much wider. The distribution of viewpoints is usually a trajectory of continuous motion, such as dolly in and out~\cite{realestate10k} or panning left and right~\cite{dl3dv}. 

To handle these issues, we design a mixed training strategy that controls the distribution of selected views, camera condition, Gaussian point clouds, and imaging depth. 

\noindent\textbf{Viewpoint Selecting.} The first step of our mixed training is to select viewpoints. For better convergence of training process, we impose two angle constraints on camera positions and orientations to ensure the noisy views and novel views have certain overlaps with the condition view. 

The first constraint is on the angle between  viewpoint positions. After normalization, this angle measures the distance of viewpoints. As the noisy views can only provide partial information, we control the angle $\theta_{cd}^{(i)}$ between the $i$-th noisy view position and the condition view position, and the angle $\theta_{dn}^{(i, j)}$ between the $i$-th noisy view position and the $j$-th novel view position. Then the constraints are
\vspace{-1.5mm}
\begin{equation}
\small
    \theta_{cd}^{(i)} \leq \theta_1,~~\theta_{dn}^{(i,j)} \leq \theta_2,
\vspace{-1.5mm}
\end{equation}
where $\theta_1$ and $\theta_2$ are hyperparameters, $1\leq i\leq N$, and $1\leq j\leq M$. The position vector can be read from the translation of camera-to-world (c2w) matrix of the viewpoint.

The second constraint is on the angle between viewpoint orientations. This angle also controls the overlap of different viewpoints. Denote the forward direction vectors of the condition view, the $i$-th noisy view, and the $j$-th novel view as $\vec{z}_{con}$, $\vec{z}_{noise}^{~(i)}$, and $\vec{z}_{nv}^{~(j)}$. Then the constraints are
\vspace{-0.7mm}
\begin{equation}
\small
    \frac{\vec{z}_{con} \cdot \vec{z}_{noise}^{~(i)}}{|\vec{z}_{con}|\cdot|\vec{z}_{noise}^{~(i)}|} \geq \text{cos}(\varphi_1),~~\frac{\vec{z}_{con} \cdot \vec{z}_{nv}^{~(j)}}{|\vec{z}_{con}|\cdot|\vec{z}_{nv}^{~(j)}|} \geq \text{cos}(\varphi_2),
\vspace{-0.6mm}
\end{equation}
Where $\varphi_1$ and $\varphi_2$ are hyperparameters. $\vec{z}$ is read from c2w.

\noindent\textbf{Reference-Point Pl\"ucker Coordinate.} To offer the camera conditions, previous methods~\cite{plucker_1, plucker_2, lgm, dmv3d, cat3d} adopt a pixel-aligned ray embedding, pl\"ucker coordinates~\cite{plucker_coordinate}, concatenated with the image as input. As shown in Fig.~\ref{fig:rppc} (a), the pixel-aligned ray embeddings are parameterized as $\bm{r} = (\bm{o} \times \bm{d}, \bm{d})$, where $\bm o$ and $\bm d$ are the position and direction of the ray landing on the pixel. Specifically, $\bm{o} \times \bm{d}$ represents the rotational effect of $\bm{o}$ relative to $\bm{d}$, showing limitations in perceiving the relative depth and  geometry.

To handle this problem, we customize a Reference-Point Pl\"ucker Coordinate (RPPC) as the camera condition. As depicted in Fig.~\ref{fig:rppc} (b), we use the point on the ray closest to the origin of the world coordinate system as the reference point to replace the moment vector, which can be formulated as
\vspace{-0.8mm}
\begin{equation}
\small
    \bm{r} = (\bm{o} - (\bm{o} \cdot \bm{d}) \bm{d}, \bm{d})
\vspace{-1.2mm}
\end{equation}
Our RPPC satisfies the translation invariance assumption of the 4D light field~\cite{4d_light_field}. Plus, compared to the moment vector, our reference point can provide more information about the ray position and the relative depth, which are beneficial for the diffusion model to capture the 3D geometry of scenes and objects. By skip connections, the reference-point information can flow through every Transformer block and the Gaussian decoder to guide the GS point cloud generation.

\noindent\textbf{Dual Gaussian Decoder.} As the depth range varies across object- and scene-level datasets, we use two MLPs to decode the Gaussian primitives for objects and scenes in mixed training. As shown in Fig.~\ref{fig:pipeline} (b), for the object-level Gaussian decoder, the nearest and farthest distances [$u_{near}$, $u_{far}$] in Eq.~\eqref{eq:distance} are set as [0.1, 4.2] and $\bm{\mu}_{t}^{(k)}$ is clipped into $[-1, 1]^3$. For the scene-level Gaussian decoder, [$u_{near}$, $u_{far}$] is set to [0, 500]. The two decoders are also controlled by the timestep embedding. In the finetuning phase, we just use a single decoder while the other is removed.

\begin{table*}[t]\vspace{-1.5mm}
	\subfloat[\small User preference and running time comparison on object generation]{\hspace{-4.5mm}
            \renewcommand{\arraystretch}{1.2}
		\label{tab:user_study}
            \setlength{\tabcolsep}{5.9pt}
		\scalebox{0.51}{\noindent
            \begin{tabular}{l c c c c c c}
                \toprule[0.15em]
                \rowcolor{color3} Method &DreamGaussian~\cite{dreamgaussian} &LGM~\cite{lgm} &DMV3D~\cite{dmv3d} &CRM~\cite{crm} &12345++~\cite{12345++} &DiffusionGS (Ours) \\
                \midrule
                User Study Score $\uparrow$ &1.94 &3.04 &3.16 &2.69 &3.81 &\bf 4.88 \\
                Runing Time (s) $\downarrow$ &120 &\bf 4.1 &31.4 &10 &60.0 & 5.8 \\
                \bottomrule[0.15em]
            \end{tabular}}}\vspace{0.5mm}
	\subfloat[\small Comparison with the SOTA 2D method PhotoNVS on~\cite{realestate10k}]{\hspace{-2mm}
            \renewcommand{\arraystretch}{1.2} 
		\label{tab:compare_2d}
            \setlength{\tabcolsep}{5.9pt} 
		\scalebox{0.51}{\noindent 
            \begin{tabular}{l c c c c c c} 
				\toprule[0.15em]
				\rowcolor{color3} Method &~~Infer Time~~ &~~Post-hoc GS Time~~ &~~PSNR $\uparrow$~~ &~~SSIM $\uparrow$~~ &~~LPIPS $\downarrow$~~ &~~FID $\downarrow$~~\\
				\midrule
				PhotoNVS~\cite{photonvs}] & 61s & 2417s &15.31 &0.5215 &0.4589 &28.30\\
				 Our DiffusionGS &\bf 6s &\bf 0s &\bf 21.63 &\bf 0.6787 &\bf 0.2743 &\bf 15.87\\
				\bottomrule[0.15em]
		\end{tabular}}}\vspace{0.5mm}
	\subfloat[\small Object generation results on ABO~\cite{abo} \label{tab:breakdown}]{\hspace{-4.5mm}
            \renewcommand{\arraystretch}{1.2}
		\label{tab:abo}
            \setlength{\tabcolsep}{5.9pt}
		\scalebox{0.6}{\noindent
            \begin{tabular}{l c c c c}
                \toprule[0.15em]
                \rowcolor{color3} Method 
                &~PSNR~$\uparrow$~  &~SSIM~$\uparrow$~ &~LPIPS~$\downarrow$~ &~FID~$\downarrow$~  \\
                \midrule
                LGM~\cite{lgm} &16.01 &0.7262 &0.3255 &86.32  \\
                GS-LRM~\cite{gslrm} &18.78 &0.7974 &0.2720 &123.55 \\
                DMV3D~\cite{dmv3d} &23.69 &0.8634 &0.1131 &32.28 \\
                \midrule
                DiffusionGS &\bf25.89 &\bf0.8880 &\bf0.0965 &\bf9.03 \\
                \bottomrule[0.15em]
            \end{tabular}}}\hspace{0mm}
	\subfloat[\small Object generation results on GSO~\cite{gso}]{ 
            \renewcommand{\arraystretch}{1.2}
		\label{tab:gso}
            \setlength{\tabcolsep}{5.9pt}
		\scalebox{0.6}{\noindent
            \begin{tabular}{l c c c c}
                \toprule[0.15em]
                \rowcolor{color3} Method &~PSNR~$\uparrow$~  &~SSIM~$\uparrow$~ &~LPIPS~$\downarrow$~ &~FID~$\downarrow$~  \\
                \midrule
                LGM~\cite{lgm} &14.27 &0.7183 &0.3003 &75.55 \\
                GS-LRM~\cite{gslrm} &17.70 &0.7950 &0.2411 &112.96 \\
                DMV3D~\cite{dmv3d} &20.82 &0.8347 &0.1289 &33.48 \\
                \midrule
                DiffusionGS &\bf 22.07 &\bf 0.8545 &\bf 0.1115 &\bf 11.52 \\
                \bottomrule[0.15em]
            \end{tabular}}}\hspace{0mm}
	\subfloat[\small Scene reconstruction on Realestate10K~\cite{realestate10k}]{\hspace{-1mm}
            \renewcommand{\arraystretch}{1.2}
		\label{tab:realestate10k}
            \setlength{\tabcolsep}{5.9pt}
		\scalebox{0.6}{\noindent
            \begin{tabular}{l c c c c c}
                \toprule[0.15em]
                \rowcolor{color3} Method &~PSNR~$\uparrow$~  &~SSIM~$\uparrow$~ &~LPIPS~$\downarrow$~ &~FID~$\downarrow$~  \\
                \midrule
                PixelNeRF~\cite{pixelnerf} &17.46 &0.5713 &0.5525 &159.52\\
                Splatter-Image~\cite{splatter_image} &18.21 &0.6115 &0.4839 &120.35 \\
                Flash3D~\cite{gslrm} &20.29 &0.6483 &0.3610 &35.03 \\
                \midrule
                DiffusionGS &\bf 21.63 &\bf 0.6787 &\bf 0.2743 &\bf 15.87 \\
                \bottomrule[0.15em]
            \end{tabular}}}\hspace{2mm}\vspace{0mm}
	\vspace{-6.5mm}
	\caption{\small User study and main quantitative results of single-view image-to-3D task on ABO~\cite{abo}, GSO~\cite{gso}, and Realestate10K~\cite{realestate10k}.}
\vspace{-4mm}
\end{table*}

\begin{figure*}[t]
	\begin{center}
		\begin{tabular}[t]{c}  \hspace{-3mm}
			\includegraphics[width=1\textwidth]{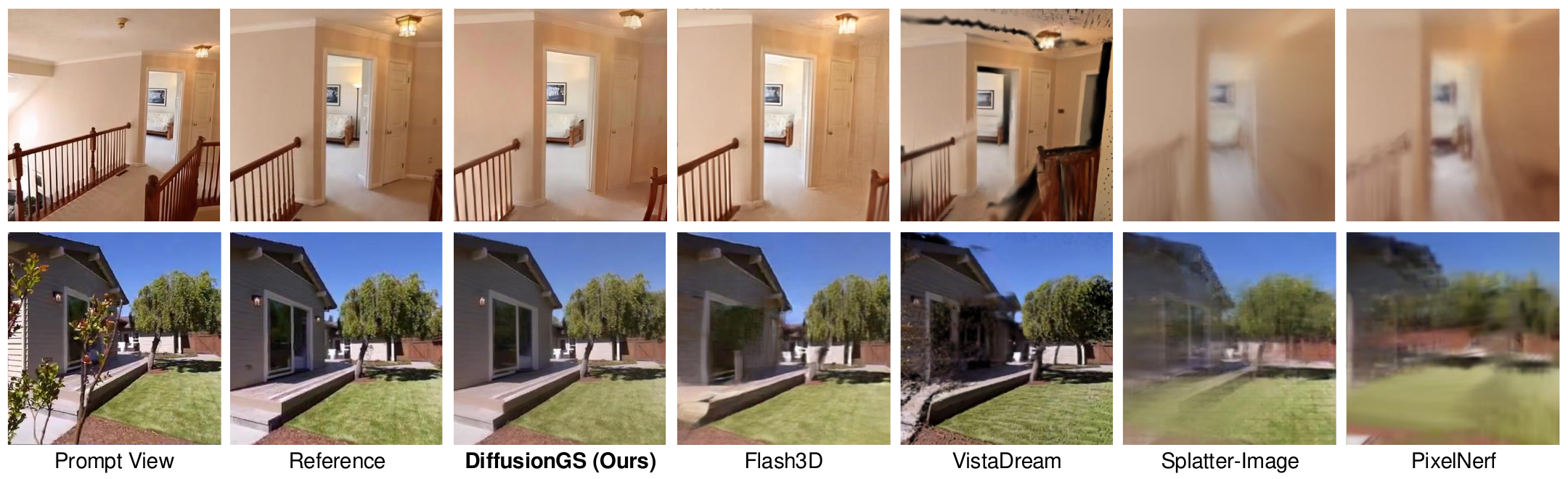}
		\end{tabular}
	\end{center}
	\vspace{-8mm}
	\caption{\small Visual results of single-view scene reconstruction. We train the feedforward methods with the same scene data for fairness. Previous methods yield blurry images or introduce artifacts.  In contrast, our method can robustly reconstruct scenes with occlusion.}
	\label{fig:scene_compare}
	\vspace{-4mm}
\end{figure*}



\noindent\textbf{Overall Training Objective.} Similar to the denoising loss $\mathcal{L}_{de}$ in Eq.~\eqref{eq:denoise_loss}, we compute $\mathcal{L}_2$ loss and perceptual loss with the same balancing hyperparameter $\lambda$ on novel views. The novel view loss is denoted as $\mathcal{L}_{nv}$. To encourage the distribution of 3D Gaussian point clouds of object-centric generation more concentrated, we design a point distribution loss $\mathcal{L}_{pd}$ for training warm-up.  $\mathcal{L}_{pd}$ is formulated as
\vspace{-1mm}
\begin{equation}
\small
    \mathcal{L}_{pd} = \underset{k}{\mathbb{E}}[l_t^{(k)} - (\frac{l_t^{(k)} - \underset{k}{\mathbb{E}}[l_t^{(k)}]}{\sqrt{\text{Var}(l_t^{(k)})}}\sigma_0 + \underset{k}{\mathbb{E}}[|\bm{o}^{(k)}|])],
\vspace{-2mm}
\label{eq:l_pd}
\end{equation}
where $\mathbb{E}$ represents the mean value, $l_t^{(k)} = |u_t^{(k)} \bm{d}^{k}|$, Var denotes the variance, and $\sigma_0$ is the target standard deviation. $\sigma_0$ is set to 0.5. Then the overall training objective $\mathcal{L}$ is
\vspace{-0.8mm}
\begin{equation}
\small
    \mathcal{L} = (\mathcal{L}_{de} + \mathcal{L}_{nv}) \cdot \mathbf{1}_{\text{iter} > \text{iter}_0} + \mathcal{L}_{pd} \cdot \mathbf{1}_{\text{iter} \leq \text{iter}_0} \cdot \mathbf{1}_{\text{object}},
\vspace{-1.2mm}
\end{equation}
where $\mathbf{1}_{\text{iter} > \text{iter}_0}$ is a conditional indicator function which equals 1 if the current training iteration (iter) is greater than the threshold ($\text{iter}_0$). $\mathbf{1}_{\text{iter} \leq \text{iter}_0}$ and $\mathbf{1}_{\text{object}}$ are similar.


\begin{figure*}[t]
	\begin{center}
		\begin{tabular}[t]{c}  \hspace{-3mm}
			\includegraphics[width=0.99\textwidth]{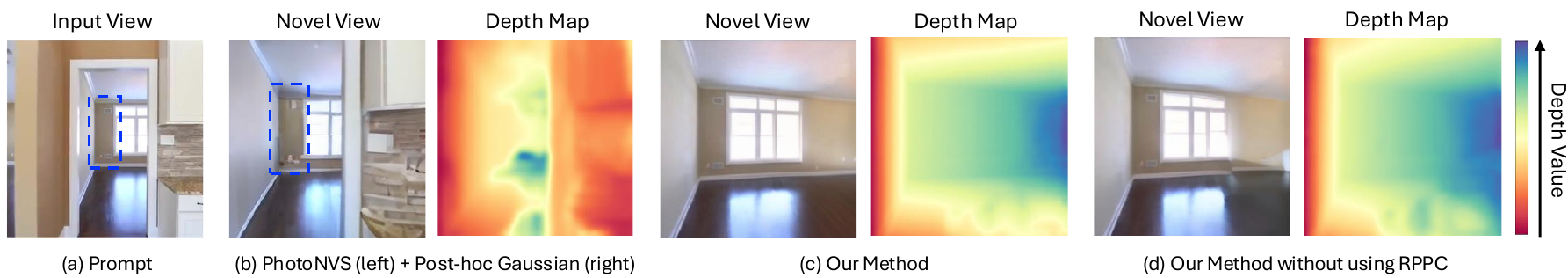}
		\end{tabular}
	\end{center}
	\vspace{-7mm}
	\caption{\small Visual comparison between the SOTA 2D method PhotoNVS~\cite{photonvs} in (b) and our method in (c) on NVS and relative depth estimation. The depth map in (b) is predicted by the post-hoc 3DGS fitting the synthesized views. (d) shows the effect of our RPPC.
    }
	\label{fig:compare_2d}
	\vspace{-3mm}
\end{figure*}

\begin{figure*}[t]
	\begin{center}
		\begin{tabular}[t]{c}  \hspace{-3mm}
			\includegraphics[width=1\textwidth]{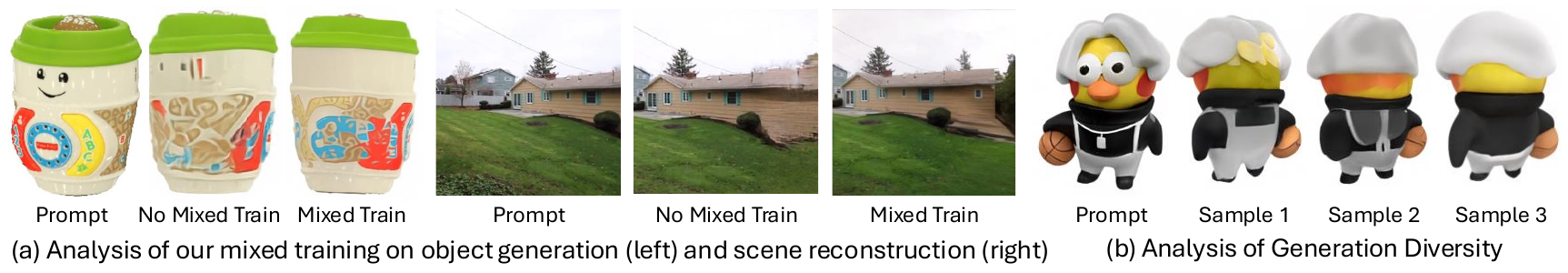}
		\end{tabular}
	\end{center}
	\vspace{-8mm}
	\caption{\small Visual analysis. (a) studies the effect of using our mixed training strategy. (b) shows different generation samples of our method.
    }
	\label{fig:visual_analysis}
	\vspace{-3mm}
\end{figure*}

\vspace{-2.5mm}
\section{Experiment}
\label{sec:experiment}
\vspace{-1.5mm}

\noindent\textbf{Dataset.} 
We use Objaverse~\cite{objaverse} and MVImgNet~\cite{mvimgnet} as the training sets for objects. We center and scale each 3D object of Objaverse into $[-1, 1]^3$, and render 32 images at random viewpoints with 50$^\circ$ FOV. For MVImgNet, we crop the object, remove the background, normalize the cameras, and center and scale the object to $[-1, 1]^3$. We preprocess 730K and 220K training samples in Objaverse and MVImgNet. We use the ABO~\cite{abo} and GSO~\cite{gso} datasets for evaluation. We adopt RealEstate10K~\cite{realestate10k} and DL3DV10K~\cite{dl3dv} as the scene-level training datasets. RealEstate10K includes 80K video clips of indoor and outdoor real scenes selected from YouTube videos. We follow the standard training/testing split. DL3DV10K contains 10510 videos of real-world scenarios, covering 96 complex categories. For all evaluation, each instance has 1 input view and 10 testing views. 

\noindent\textbf{Implementation Details.} We implement DiffusionGS by Pytorch~\cite{pytorch} and train it with Adam optimizer~\cite{adam}. To save GPU memory, we adopt mixed-precision training~\cite{amp} with BF16, sublinear memory training~\cite{sublinear}, and deferred GS rendering~\cite{arf}. In mixed training, we use 32 A100 GPUs to train the model on Objaverse, MVImgNet, RealEstate10K, and DL3DV10K for 40K iterations at the per-GPU batch size of 16. Then we finetune the model on the object- and scene-level datasets with 64 A100 GPUs for 80K and 54K iterations at the per-GPU batch size of 8 and 16. The learning rate is linearly warmed up to 4e$^{-4}$ with 2K iterations and decays to 0 using cosine annealing scheme~\cite{cosine}. Finally, we scale up the training resolution from 256$\times$256 to 512$\times$512 and finetune the model for 20K iterations. 

\vspace{-1mm}
\subsection{Comparison with State-of-the-art Methods}
\vspace{-1mm}
\noindent\textbf{Single-view Object Generation.} Fig.~\ref{fig:object_compare} shows the visual comparison of object-level generation on ABO, GSO, real-camera image~\cite{openillu}, and text-to-image prompted by FLUX. We compare our method with five state-of-the-art (SOTA) methods including a one-stage 3D diffusion DMV3D~\cite{dmv3d}, three 2D multi-view diffusion-based methods (LGM~\cite{lgm}, CRM~\cite{crm}, and 12345++~\cite{12345++}), and an SDS-based method DreamGS~\cite{dreamgaussian}. Previous methods render over-smoothed images or distort  3D geometry. In contrast, our method robustly generates clearer novel views and perfect 3D geometry with prompt views of any directions, while preserving fine-grained details. Even when the front view, which previous methods specialize in, is given (third and fourth row), our method still yields better view consistency by retaining the face details of the dolls. While the methods based on 2D multi-view diffusion introduce cracks, artifacts, and blur to the faces when ``stitching" unaligned multi-view images.

We conduct a user study by inviting 25 people to score the visual quality of the generation results of 14 objects according to the 3D geometry, texture quality, and alignment with the prompt view. The user study score ranges from 1 (worst) to 6 (best). 
Tab.~\ref{tab:user_study} reports the results and running time at the size of 256$\times$256. Our method achieves the highest score while enjoying over 5$\times$ and 10$\times$ inference speed compared to the recent best 3D diffusion DMV3D and multi-view diffusion-based method 12345++. Tab.~\ref{tab:abo} and~\ref{tab:gso} list the results of object generation on the ABO and GSO datasets. DiffusionGS surpasses DMV3D by 2.2/1.25 dB in PSNR and 23.25/21.96 in FID score on ABO/GSO.

We chain stable diffusion~\cite{stable_diffusion} or FLUX with DiffusionGS to perform text-to-3D in Fig.~\ref{fig:teaser} and \ref{fig:wild_object_demo}, our method can handle hard cases with furry appearance, shadow, flat illustration, complex geometry, and even specularity.

\begin{table}
	\begin{center}
            \renewcommand{\arraystretch}{1}
            \setlength{\tabcolsep}{5.9pt}
		\scalebox{0.65}{\noindent
            \begin{tabular}{l c c c c c}
                \toprule[0.15em]
                \rowcolor{color3} Method &~Baseline~ &~+ Our Diffusion~ &~+ $\mathcal{L}_{pd}$~ &~+ Mixed Training~ &~+ RPPC~  \\
                \midrule
                PSNR $\uparrow$ &17.63 &20.57 &20.94 &21.73 &22.07\\
                SSIM $\uparrow$ &0.7928 &0.8120 &0.8423 &0.8515 &0.8545 \\
                LPIPS $\downarrow$ &0.2452 &0.1417 &0.1218 & 0.1196 &0.1115\\
                FID $\downarrow$ &118.31 &47.86 &28.41 &17.79 &11.52\\
                \bottomrule[0.15em]
            \end{tabular}}
            \vspace{-2mm}
            \caption{Ablation study. Results on the GSO~\cite{gso} dataset are listed.}
            \label{tab:ablation_new}
            \vspace{-9mm}
        \end{center}
\end{table}

\noindent\textbf{Single-view Scene Reconstruction.} We compare our method with three feedforwad methods (Flash3D~\cite{flash3d}, Splatter-Image~\cite{splatter_image}, and pixelNeRF~\cite{pixelnerf}) and one SDS-based method VistaDream~\cite{vistadream}. For fair comparison, we train the feedforward methods with the same scene data as DiffusionGS. Tab.~\ref{tab:realestate10k} reports the results on RealEstate10K. Our method outperforms the SOTA method Flash3D~\cite{gslrm} by 1.34 dB in PSNR and 19.16 in FID score. Fig.~\ref{fig:teaser},~\ref{fig:scene_demo}, and~\ref{fig:scene_compare} depict the visual results of indoor and outdoor scene reconstruction. In Fig.~\ref{fig:scene_compare}, pixelNeRF and Splatter-Image render blurry images. Although using monocular depth estimator, Flash3D and VistaDream still produce blur, artifacts, noise, and black spots in the occluded region of novel views. In contrast, as our DiffusionGS can generate the views along the camera trajectory to predict more refined and structured Gaussian point clouds, it can reconstruct clearer details of scenes with occlusion, \textbf{without using depth estimator}.

We also compare our method and the SOTA 2D view synthesis method PhotoNVS~\cite{photonvs} with post-hoc Gaussian fitting in Tab.~\ref{tab:compare_2d}. Our method outperforms PhotoNVS by 6.32 dB in PSNR and 12.43 in FID. Fig.~\ref{fig:compare_2d} (b) and (c) show the comparison of rendered views and depth maps. Due to the lack of 3D models, PhotoNVS show limitations in perceiving 3D structure and cannot ensure view-consistency. In Fig.~\ref{fig:compare_2d} (b), PhotoNVS fails to reveal the room. The generated images also contain many view-inconsistent contents, as depicted in the blue boxes of Fig.~\ref{fig:compare_2d} (a) and (b). Plus, using post-hoc 3D Gaussians to fit the generated views takes a long time ($\sim$ 40 minutes) and suffers from the over-fitting issue due to view inconsistency. Thus, in Fig.~\ref{fig:compare_2d} (b), the post-hoc Gaussian fails to predict the depth. \textbf{In contrast}, as shown in Fig.~\ref{fig:compare_2d} (c), our method can 
generate more view-consistent images and more accurate 3D structure in 6s.

We chain Sora~\cite{sora} with our method to perform text-to-scene in Fig.~\ref{fig:teaser}. Our method can reliably render novel views for both indoor and outdoor scenes prompted by Sora.

\vspace{-0.7mm}
\subsection{Ablation Study}
\vspace{-0.8mm}
\noindent\textbf{Break-down Ablation.} To study the effect of each component towards higher performance, we adopt the denoiser without timestep control as the baseline to conduct a break-down ablation. We train it on the object-level datasets with a single-view input and the same amount of supervised views as DiffusionGS.  Results on GSO~\cite{gso} are reported in Tab.~\ref{tab:ablation_new}. The baseline yields poor results of 17.63 dB in PSNR and 118.31 in FID. When applying our diffusion framework, loss $\mathcal{L}_{pd}$ in Eq.\eqref{eq:l_pd}, scene-object mixed training without RPPC, and RPPC, the model gains by 2.94, 0.37, 0.79, 0.34 dB in PSNR and drops by 70.45, 19.45, 10.62, 6.27 in FID.


\noindent\textbf{Analysis of Mixed Training.} We conduct an analysis of our scene-object mixed training in Fig.~\ref{fig:visual_analysis} (a). For fair comparison, models are trained with the same iterations whether with or without mixed training. \textbf{(i)} The left part shows the effect on object generation. After using the mixed training, the textures of the cup become clearer and more realistic, and the artifacts on the back are reduced. \textbf{(ii)} The right part depicts the effect on scene reconstruction. When applying our mixed training, DiffusionGS can better capture the 3D geometry and reconstruct the house with less artifacts. The improvement of our mixed training for scene reconstruction on Realestate10K is 0.61 dB in PSNR and 10.53 in FID.

\noindent\textbf{Analysis of RPPC.} \textbf{(i)} Using RPPC leads to a quantitative improvement of 0.28 dB in PSNR and 7.09 in FID for scene reconstruction on Realestate10K. \textbf{(ii)} Fig.~\ref{fig:compare_2d} (c) and (d) analyze the effect of RPPC. Using RPPC can render more accurate scene structures (window, floor, \emph{etc.}) and depth. 

\noindent\textbf{Analysis of Generation Diversity.} We change the random seed with the same prompt view in Fig.~\ref{fig:visual_analysis} (b). Our method can generate different shapes and textures for 3D assets.

\vspace{-2.2mm}
\section{Conclusion}
\vspace{-1.8mm}
\label{sec:conclusion}
In this paper, we propose a novel framework, DiffusionGS, for 3D object generation and scene reconstruction from a single view. Our DiffusionGS directly outputs 3D Gaussian point clouds at each timestep to enforce view consistency and only requires 2D renderings for supervision. In addition, we develop a scene-object mixed training strategy with a new camera conditioning method RPPC to learn a general prior capturing better 3D geometry and texture representations. Experiments show that our method outperforms SOTA methods while enjoying a faster speed of 6 seconds.

{
    \small
    \bibliographystyle{ieeenat_fullname}
    \bibliography{main}
}

\end{document}